\documentclass[11pt,a4paper]{article}

\usepackage[utf8]{inputenc}
\usepackage[T1]{fontenc}
\usepackage[english]{babel}
\usepackage{lmodern}
\usepackage{microtype}
\usepackage[a4paper,margin=2.5cm]{geometry}
\usepackage{amsmath,amssymb}
\usepackage{graphicx}
\graphicspath{{pic/}{./}}
\usepackage{booktabs,tabularx,array}
\usepackage{authblk}
\usepackage{caption}
\usepackage{float}
\usepackage[section]{placeins}
\usepackage[numbers,sort&compress]{natbib}
\usepackage[hidelinks]{hyperref}
\usepackage{url}

\newcolumntype{L}{>{\raggedright\arraybackslash}X}
\newcolumntype{C}{>{\centering\arraybackslash}X}
\newcommand{\email}[1]{\href{mailto:#1}{\nolinkurl{#1}}}
\newcommand{\doi}[1]{\href{https://doi.org/#1}{doi:\nolinkurl{#1}}}
\newcommand{\keywords}[1]{%
  \par\smallskip\noindent\textbf{Keywords:} #1\par\medskip}
\title{\bfseries Interpreting Supervised Pairwise Graph Representations:\\
Local Evidence, Confounding, and Provable Post-hoc Correction}

\author[1]{Ilya Brekin}
\author[2,*]{Ariela Moldavskaya}
\author[3]{Elina Sidorova}
\author[4]{Ural Valeev}
\affil[1]{HSE University, Moscow, Russia; \email{brekinilya@gmail.com}}
\affil[2]{Technion -- Israel Institute of Technology, Technion City,
Haifa 3200003, Israel}
\affil[3]{Lomonosov Moscow State University, Moscow, Russia;
\email{elina_sidorova@internet.ru}}
\affil[4]{Ufa University of Science and Technology, Ufa, Russia;
\email{valeev.ural@mail.ru}}
\affil[*]{Corresponding author. E-mail address:
\email{arielam@campus.technion.ac.il}}
\date{}

\begin{document}
\maketitle

\begin{abstract}
Supervised pairwise graph representations convert pairwise predictive relations
into sample-specific graphs, but graph-level accuracy does not by itself show
that higher-order topology carries the signal. In a controlled Transformer
error-detection task, supervised pairwise graphs outperform
attention-as-adjacency (ROC-AUC $0.839$ versus $0.559$; the comparable
no-confidence pair is $0.687$ versus $0.559$), but their apparent advantage
over model confidence disappears after controlling for sequence length, which
is almost perfectly recoverable from the pooled hidden state ($R^2=0.9995$).
On published HCP Emotion ensemble graphs, signed node aggregation already acts
as an ensemble of supervised pairwise predictions. Edge magnitude alone is near
chance; topology-aware augmentations change ROC-AUC by at most $0.004$, with
only sign-separated degree showing small positive changes across all tested
graph sizes, while a validation-tuned three-layer GCN provides no measurable
improvement over the signed node-summary baseline. A Fisher post-hoc corrector
satisfies its distribution-agnostic guarantees across the evaluated feature
representations and remains stable under reduced labelled budgets; in a
near-error-free fMRI calibration regime the bound correctly becomes
statistically uninformative. A separate cross-attention analysis identifies a
redundant reversal circuit and downstream EOS-timing and copy failures.
Higher-order topology should therefore be credited only after local supervised
evidence and recoverable nuisance variables have been excluded.
\end{abstract}


\par\medskip

\section{Introduction}
Supervised pairwise graph representations turn high-dimensional observations
into sample-specific graphs whose edges are produced by pairwise predictors.
This creates an attribution problem: a successful graph model may exploit
supervised information already present in individual edges, local node
summaries, or nuisance variables rather than higher-order topology.

We study this problem in two deliberately different regimes. Setup~1 is a
controlled sequence-reversal Transformer in which nodes are anonymous latent
coordinates and the target is prediction correctness. Setup~2 uses published
HCP Emotion ensemble graphs, where nodes are named brain regions and signed
edges are held-out pairwise evidence for cognitive state. We adopt an
edge $\rightarrow$ node $\rightarrow$ higher-order information hierarchy as a
standard of evidence and add nuisance-variable controls before accepting a
topological interpretation.

The team's work extends beyond applying this criterion. We construct the
controlled Transformer benchmark and its supervised pairwise graph
representation, introduce a direction-invariant decisiveness edge score for
anonymous latent coordinates, implement an independent subject-safe raw-HCP
reconstruction path, perform topology and dimensional GCN stress tests on
published HCP ensemble graphs, evaluate Fisher post-hoc correction under
reduced labelled budgets, and conduct a separate mechanistic analysis of
Transformer cross-attention.

The resulting picture is asymmetric but coherent. In the Transformer regime,
the graph representation is more informative than raw attention, yet its
apparent edge over confidence is explained by sequence length. In HCP Emotion,
supervised edge sign already carries nearly all measurable discriminative
information, leaving little room for the evaluated higher-order descriptors or
GCN. The corrector and cross-attention analyses provide two additional positive
results independent of the topology claim.

\section{Related Work}
Synolitic Graph Neural Networks (SGNNs) were introduced as a general framework
for constructing sample-specific supervised pairwise graphs by Zaikin et
al.~\cite{zaikin2026synolitic}. Related synolitic constructions were
demonstrated on clinical and physiological data by Krivonosov et
al.~\cite{krivonosov2025}. Vlasenko et al.~specialize the approach to HCP fMRI
ensemble graphs~\cite{vlasenko2025fmri}. Their published accuracy range for
logistic regression on mean incident edge weights is $97.07$--$99.74\%$,
while the separately reported GNN range on ensemble graphs is
$88.00$--$99.42\%$. These ranges are not paired task-by-task comparisons. For
the Emotion paradigm studied here, however, the published values are directly
comparable: $98.45\%$ accuracy for logistic regression on mean incident edge
weights versus $88.00\%$ for the GNN, a $10.45$ percentage-point gap.

Tyukin et al.~develop weakly supervised post-hoc correctors with
distribution-agnostic, non-asymptotic guarantees~\cite{tyukin2024coping,tyukin2024weakly}.
We use this framework only as a downstream decision layer: validity of the
correction bound is not treated as evidence that a graph representation is
necessary.

Attention weights are often inspected as model structure, but Jain and Wallace
\cite{jain2019} and Wiegreffe and Pinter~\cite{wiegreffe2019} caution against
treating attention as explanation without an explicit test. Accordingly,
attention-as-adjacency is only a direct structural baseline here; the separate
mechanistic analysis tests a concrete mirror-position hypothesis in
cross-attention.

Our attribution logic is related to control-task/selectivity ideas in
representation probing (Hewitt and Liang~\cite{hewitt2019}) and nuisance control
in neuroimaging decoding (Snoek et al.~\cite{snoek2019}). We do not claim these
diagnostics as new. The contribution is their systematic use in supervised
pairwise graphs, where target information can enter through edge construction
itself.

\section{Model Description}
\label{sec:model}
For samples $x^{(n)}\in\mathbb{R}^d$ with binary labels $y^{(n)}$, a
probabilistic classifier is associated with each unordered coordinate pair
$(i,j)$. Its sample-specific output defines a weighted graph
$G^{(n)}=(V,E,W^{(n)})$. Pairwise classifiers are trained within Setup~1 and
inherited from the published Vlasenko construction in Setup~2.

For anonymous Transformer coordinates we remove arbitrary class polarity and
use the direction-invariant decisiveness score introduced here as an
experimental instantiation, not as a proposed replacement for signed pairwise
constructions:
\[
w_{ij}=2\left|q_{ij}-\frac12\right|.
\]
For named HCP brain regions we preserve the published signed evidence
\[
w_{ij}=P(c_2\mid i,j)-P(c_1\mid i,j).
\]
The difference between these edge definitions follows the semantics of the
nodes rather than representing two competing graph constructions. In Setup~1,
latent coordinates are anonymous: exchanging the arbitrary class orientation
of a pairwise classifier should not change the relation represented by an edge,
which motivates the direction-invariant decisiveness score. In Setup~2, nodes
are named brain regions and the published edge sign explicitly encodes evidence
for one cognitive state versus the other; discarding the sign would therefore
remove supervised information rather than eliminate an arbitrary orientation.

We distinguish individual edge evidence, local node summaries, and higher-order
graph processing. The principal signed HCP summary is
\[
\bar w_i=\frac{1}{d-1}\sum_{j\ne i}w_{ij}.
\]
When edges are supervised predictions of the target, this local aggregation
already behaves as an ensemble of weak predictors. We operationalize
attribution sequentially: first test whether predictive performance is
explained by individual edge information or recoverable nuisance variables;
next compare against local node summaries; only then test whether topology-aware
descriptors or message passing add measurable predictive information. Detailed
evaluation procedures are given in Section~\ref{sec:protocol}.

Node identity is fixed in both settings, so the principal graph learners use
identity-preserving flatten readout. Setup~1 uses GATv2~\cite{brody2022gatv2};
Setup~2 uses a three-layer GCN backbone~\cite{kipf2017gcn} for comparability
with the published ensemble-graph study. Both internally constructed
Transformer graphs and externally supplied HCP ensemble graphs are converted to
the same validated downstream graph interface. This separates graph
construction from graph learning and allows the same attribution logic to be
applied without implying that the two edge generators have the same provenance
or semantics.

\section{Datasets and Preprocessing}
\label{sec:data}

\subsection{Controlled Transformer benchmark}
Setup~1 contains $10{,}000$ sequence-reversal examples from a Transformer
\cite{vaswani2017attention} with two encoder and two decoder layers, four
attention heads, $d_{\rm model}=64$, and approximately $236$K parameters.
Training is stopped at $72.4\%$ exact-match accuracy so that held-out data
retain a substantial mixture of correct and erroneous predictions. Mean-pooled
encoder hidden states provide the $64$-dimensional representation. The fixed
split is $60/20/20$ for training, calibration, and test; $10\%$ of training is
reserved for GNN selection.

\subsection{Published HCP ensemble graphs}
Setup~2 uses published HCP Emotion ensemble graphs~\cite{vlasenko2025fmri} for
fear-versus-neutral classification, based on Human Connectome Project data
\cite{elam2021,vanessen2013}: $581$ subjects, right-to-left phase encoding,
Glasser parcellation, and up to $d=379$ named brain-region nodes. Both states
from a subject remain in the same partition: $349/116/116$ subjects
($698/232/232$ graphs) for training/calibration/test. Each published graph is
held out with respect to its own subject. Because our partition is imposed after the original held-out folds were
generated, a graph assigned to our test split may have been produced by an
ensemble that observed subjects we place in training. This can inflate
\emph{absolute} AUC values, and we therefore do not interpret them as
independent estimates of decoding accuracy. Our comparisons are paired: every model consumes identical fixed graphs
under an identical partition. They are nevertheless conditional on this
precomputed bundle and do not provide an independent assessment of the full
graph-generation and decoding pipeline.

\subsection{Independent raw-HCP reconstruction}
In parallel, we implemented an independent raw-HCP path with fixed-length
preprocessing, subject-safe splitting, out-of-fold pairwise graph generation,
signed edge preservation, and a common PyTorch Geometric interface. It is used
for preprocessing audits rather than the principal comparative HCP results;
the recording-length audit is reported in Section~\ref{sec:raw-audit},
and evaluation procedures are described in Section~\ref{sec:protocol}.

\section{Experimental Protocol}
\label{sec:protocol}

\subsection{Metrics and baseline comparisons}
Held-out ROC-AUC is the primary discrimination metric; exact-match accuracy is
used only to characterize the upstream Transformer and the $L=20$ error regime.
Paired comparisons use the same held-out samples or fixed graph realizations.

Setup~1 compares confidence, raw hidden states, flattened graph features,
raw-attention graphs, and synolitic GNNs. Setup~2 compares signed node-summary
logistic regression, topology-aware augmentations, and a validation-tuned GCN.

\subsection{Data separation and calibration}
In Setup~1, pairwise classifiers are fitted on training data only;
$10\%$ of training is reserved for GNN early stopping and model selection,
and the calibration split is reserved for the Fisher corrector.
In Setup~2, GCN hyperparameters are selected separately for each $d$ and
compared with LogReg on the same fixed graph realizations. The distinction
between the published graph bundle and the independent raw-HCP reconstruction
is maintained throughout.

\subsection{Transformer nuisance controls}
Exact-length strata are used for the nuisance-controlled
GNN-versus-confidence comparison; uncertainty is obtained by bootstrap
resampling preserving the strata. The bootstrap resample count for this
ROC-AUC analysis was not preserved in the available experimental record.
It is distinct from the mirror-fidelity bootstrap described below.

\subsection{HCP region sampling and GCN selection}
\label{sec:gcn-protocol}
HCP topology probes are evaluated at $d=34,50,100$ over three random region
subsets. The GCN ladder uses $d=34,50,100,200$ over five random subsets and
$d=379$ once on the complete graph; the two robustness experiments are not
numerically compared as if they shared identical draws. Signed and
magnitude-only edge information are evaluated separately.

For each graph dimensionality, the three-layer GCN was tuned separately
on the validation split. Table~\ref{tab:gcn_tune} reports the searched
and validation-selected hidden dimensions. Configuration selection uses a
single region draw while evaluation uses five, so selection is not strictly
nested. This may favour the tuned GCN over the untuned logistic baseline
and limits interpretation of the reported differences.

\begin{table}[H]
\centering
\small
\caption{Validation-selected GCN dimensions.}
\label{tab:gcn_tune}
\begin{tabularx}{\linewidth}{@{}cLLcc@{}}
\toprule
$d$ & $h$ searched & $h'$ searched & $h$ chosen & $h'$ chosen\\
\midrule
34  & 8,12,16       & 8,12      & 8  & 8\\
50  & 12,16,24,32   & 8,16      & 12 & 8\\
100 & 16,24,32,48   & 8,16,24   & 48 & 16\\
200 & 24,32,48      & 8,16,24   & 32 & 16\\
\bottomrule
\end{tabularx}
\end{table}

\subsection{Corrector and mechanistic evaluation}
The Fisher corrector is evaluated across feature representations and labelled
correction budgets, including $25$ examples. Cross-attention mirror fidelity is
compared within length strata using a stratified bootstrap with $2{,}000$
resamples. Error analysis separately tests intact mirror alignment and, at
$L=20$, EOS timing versus conditional content-copy failure.

\section{Results}

\subsection{Pairwise construction versus attention-as-adjacency}
\label{sec:pairwise}
The raw-attention graph is a weak error detector (ROC-AUC $0.559$), whereas the
strongest synolitic configuration reaches $0.839$ (Table~\ref{tab:main1}).
Because the latter also receives model confidence, the more comparable
no-confidence synolitic model reaches $0.687$ versus $0.559$ for raw attention.
However, $0.687$ is essentially identical to logistic regression on the
original hidden representation ($0.686$), and logistic regression on flattened
graph features reaches $0.699$. Thus supervised pairwise construction is more
informative than direct attention-as-adjacency, but these comparisons do not
establish a gain over the underlying hidden representation or attribute the
full $0.839$--$0.559$ gap solely to topology.

\begin{table}[!tbh]
\centering
\caption{Setup~1 principal test ROC-AUC results ($N_{\rm test}=2000$; base exact-match accuracy $72.4\%$).}
\label{tab:main1}
\small
\begin{tabularx}{\linewidth}{@{}Lcc@{}}
\toprule
Method & Confidence & ROC-AUC\\
\midrule
Softmax confidence & --- & 0.799\\
LogReg on raw hidden states & No & 0.686\\
LogReg on flat graph features & No & 0.699\\
LogReg on graph features + confidence & Yes & 0.841\\
GNN on raw-attention graphs & No & 0.559\\
No-confidence synolitic GNN & No & 0.687\\
Synolitic GNN ensemble & Yes & 0.839\\
\bottomrule
\end{tabularx}
\end{table}

\subsubsection{Readout and architecture ablations}

The evaluated configurations do not form a complete factorial design:
a mean-pool GNN without confidence was not evaluated. Table~\ref{tab:readout}
therefore lists only configurations actually run.

\begin{table}[H]
\centering
\small
\caption{Readout and architecture ablations in Setup~1.}
\label{tab:readout}
\begin{tabularx}{\linewidth}{@{}LLcc@{}}
\toprule
Model / representation & Readout & Confidence & ROC-AUC\\
\midrule
LogReg on flattened graph features
    & Flatten & No & 0.699\\
LogReg on mean/max-pooled graph features
    & Mean/max pooling & No & 0.634\\
No-confidence synolitic GNN
    & Flatten & No & 0.687\\
Synolitic GNN ensemble
    & Flatten & Yes & 0.839\\
Mean-pool GNN
    & Mean pooling & Yes & 0.834\\
LogReg on graph features + confidence
    & Flatten & Yes & 0.841\\
\bottomrule
\end{tabularx}
\end{table}

\subsubsection{Exploratory graph-statistic diagnostics}

We compare several graph-derived summary statistics between correct and
erroneous Transformer predictions. The observed differences motivated the
stronger attribution controls used in the main analysis; they are not
interpreted as evidence that higher-order topology causes the effect.

\begin{table}[H]
\centering
\small
\caption{Correct versus erroneous Transformer predictions.}
\label{tab:graph_statistics}
\begin{tabularx}{\linewidth}{@{}Lrrrr@{}}
\toprule
Statistic & Correct & Error & Difference & $p$-value\\
\midrule
Mean strength
    & 4.376 & 4.131 & +0.246 & $6.7\times10^{-15}$\\
Max strength
    & 30.14 & 25.97 & +4.17 & $3.0\times10^{-18}$\\
Mean closeness
    & 0.242 & 0.199 & +0.043 & $7.3\times10^{-20}$\\
Mean betweenness
    & 0.0103 & 0.0084 & +0.0019 & $3.6\times10^{-21}$\\
Mean edge weight
    & 0.778 & 0.762 & +0.016 & $6.7\times10^{-15}$\\
Number of decisive edges
    & 168.8 & 105.3 & +63.4 & $1.2\times10^{-10}$\\
\bottomrule
\end{tabularx}
\end{table}

\subsubsection{Node-feature ablation}

To test whether the Setup~1 signal can be attributed to any single
graph-derived node statistic, we fit logistic regression to each node feature
separately and compare it with the original hidden representation.
No individual graph-derived statistic (degree, strength, closeness, or
betweenness) exceeds the raw-hidden-state baseline. The value coordinate is
not a graph-derived statistic and is reported separately for completeness.
The ablation therefore does not identify a single graph-derived node statistic
that explains the Setup~1 result; it supports the stronger nuisance-controlled
and representation-level comparisons in Section~\ref{sec:length}.

\begin{table}[H]
\centering
\small
\caption{Setup~1 node-feature ablation.}
\label{tab:node_features}
\begin{tabularx}{\linewidth}{@{}Lc@{}}
\toprule
Feature & Test ROC-AUC\\
\midrule
value (raw hidden-state coordinate) & 0.714\\
degree & 0.670\\
strength & 0.674\\
closeness & 0.649\\
betweenness & 0.629\\
all five features (flat) & 0.696\\
raw hidden states (baseline) & 0.686\\
\bottomrule
\end{tabularx}
\end{table}

\subsection{Sequence length as a Transformer nuisance variable}
\label{sec:length}
In Setup~1, adding sequence length to confidence raises ROC-AUC to $0.840$,
essentially matching the pooled synolitic GNN result of $0.839$
(Table~\ref{tab:length}). Length is almost perfectly recoverable from the
pooled hidden state (ridge $R^2=0.9995$). Within exact-length strata, the
ranking reverses: GNN $0.806$ versus confidence $0.817$, with bootstrap $95\%$
CI $[-0.018,-0.003]$. Thus the apparent pooled advantage over confidence does
not survive nuisance control. Together with the no-confidence comparisons in
Section~\ref{sec:pairwise}, these results provide no clear evidence of a topology-specific
advantage over simpler representations; the no-confidence GNN may nevertheless
be suboptimal, so this is not an impossibility claim.

\begin{table}[H]
\centering
\small
\caption{Pooled Transformer comparisons.}
\label{tab:length}
\begin{tabularx}{\linewidth}{@{}Lc@{}}
\toprule
Predictor & Test ROC-AUC\\
\midrule
Length only, LogReg & 0.628\\
Softmax confidence & 0.799\\
Confidence + length, LogReg & 0.840\\
Synolitic GNN ensemble & 0.839\\
\bottomrule
\end{tabularx}
\end{table}

\subsection{Local evidence and topology in HCP}
\label{sec:hcp-locality}

\subsubsection{Raw-series record-length audit}
\label{sec:raw-audit}
\begin{table}[!tbh]\centering\small
\caption{ROC-AUC of trivial structural predictors.}\label{tab:fmri_confound}
\begin{tabularx}{\linewidth}{@{}Lccc@{}}\toprule Paradigm & Duration & No. blocks & After cropping\\\midrule
Emotion & 0.941 & 0.500 & 0.500\\
Social & 0.774 & 0.528 & 0.500\\
Motor & 0.500 & 0.500 & 0.500\\
Language & 0.848 & 1.000 & 1.000 (excluded)\\\bottomrule
\end{tabularx}\end{table}
Emotion and Social lose duration-related predictability after deterministic,
class-independent fixed-length cropping; Motor shows no measurable confound;
Language remains perfectly predictable from block count and is excluded. This
audit applies only to the reconstructed raw-HCP path.

\subsubsection{Signed node summaries and topology-aware augmentations}

The HCP result has a different mechanism. Each edge is signed supervised
evidence, $w_{ij}=P(c_2\mid i,j)-P(c_1\mid i,j)$, so signed node aggregation
already ensembles pairwise predictions. Edge magnitude alone remains near
chance (ROC-AUC $\approx0.47$--$0.54$). Across topology probes
(Table~\ref{tab:topology}), changes are concentrated near zero: the largest
individual positive deviation is $+0.004$ at $d=50$, while sign-separated degree
is the only augmentation positive at all tested sizes ($+0.003,+0.003,+0.001$
for $d=34,50,100$).

\begin{table}[!tbh]\centering\small
\caption{$\Delta$ROC-AUC relative to signed mean-incident-weight LogReg; mean of three random region draws.}\label{tab:topology}
\begin{tabularx}{\linewidth}{@{}Lccc@{}}\toprule
Added features & $d=34$ & $d=50$ & $d=100$\\\midrule
Magnitude centralities, dense & $-0.000$ & $-0.002$ & $-0.000$\\
Magnitude centralities, top 20\% & $-0.001$ & $-0.002$ & $-0.000$\\
Magnitude centralities, top 10\% & $+0.000$ & $-0.002$ & $+0.001$\\
Sign-aware centralities & $-0.002$ & $+0.004$ & $-0.000$\\
Sign-separated degree & $+0.003$ & $+0.003$ & $+0.001$\\\bottomrule
\end{tabularx}\end{table}

\subsubsection{Dimensional scaling}

\begin{figure}[H]
\centering
\includegraphics[width=0.98\linewidth]{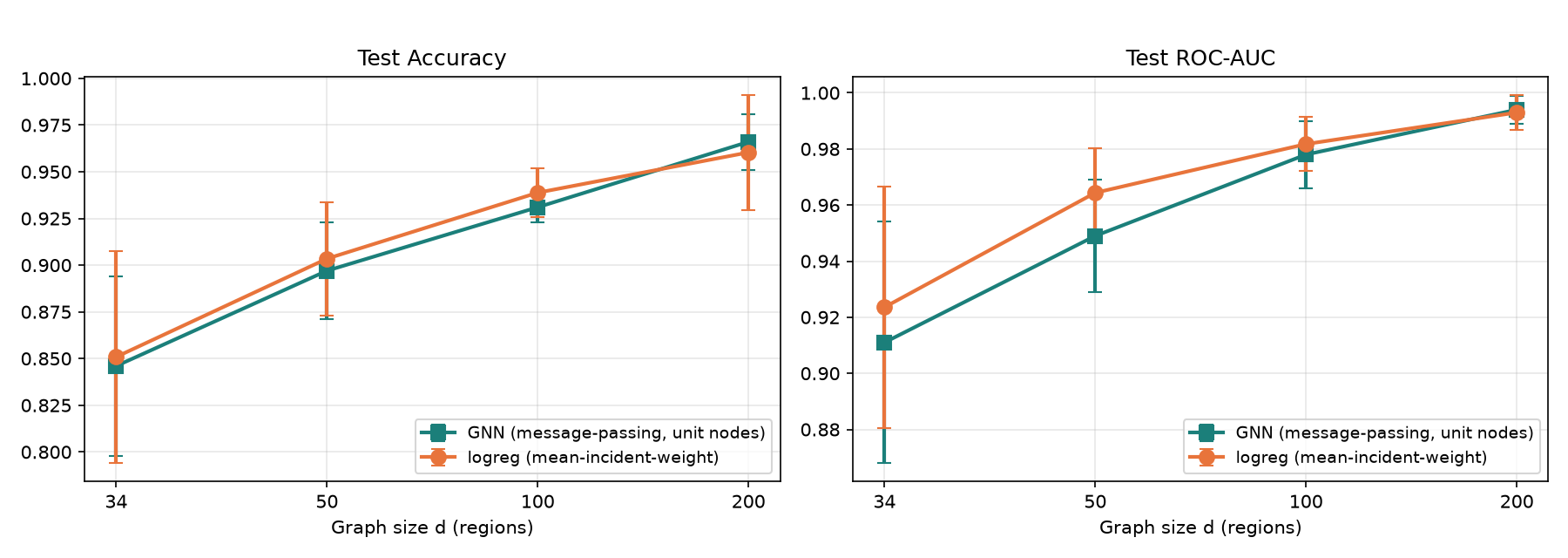}
\caption{Setup~2 dimensionality ladder. For
$d\in\{34,50,100,200\}$, points summarize repeated random region
subsets; the complete-graph result at $d=379$ is reported in the text and is not
shown in this figure.}
\label{fig:setup2_scaling}
\end{figure}

Figure~\ref{fig:setup2_scaling} compares the signed node-summary baseline
with the validation-tuned GCN across the sampled graph sizes described in
Section~\ref{sec:gcn-protocol}.

For this construction and the tested GCN family, a validation-tuned three-layer
GCN provides no measurable improvement over the signed node-summary baseline.
LogReg/GCN ROC-AUC pairs are $0.924/0.911$, $0.964/0.949$, $0.982/0.978$,
$0.993/0.994$, and $0.999/0.996$ for $d=34,50,100,200,379$. At small $d$,
where headroom remains, the local baseline is slightly stronger; at $d=200$
and $379$ both models are near ceiling. Configuration selection and evaluation
share region draws, so the reported GCN values may carry a small optimistic
bias. The non-nested selection described in Section~\ref{sec:gcn-protocol}
may favour the tuned GCN over the untuned logistic baseline; these comparisons
are therefore interpreted conditional on the evaluated configurations.

The local-aggregation mechanism helps explain the specific $10.45$
percentage-point Emotion gap reported by Vlasenko et
al.~\cite{vlasenko2025fmri}: the local statistic already aggregates supervised
target evidence, leaving little measurable additional information for the
tested higher-order graph processing.

\subsection{Provable post-hoc correction}
\label{sec:corrector}
\begin{figure}[H]
\centering
\includegraphics[width=0.94\linewidth]{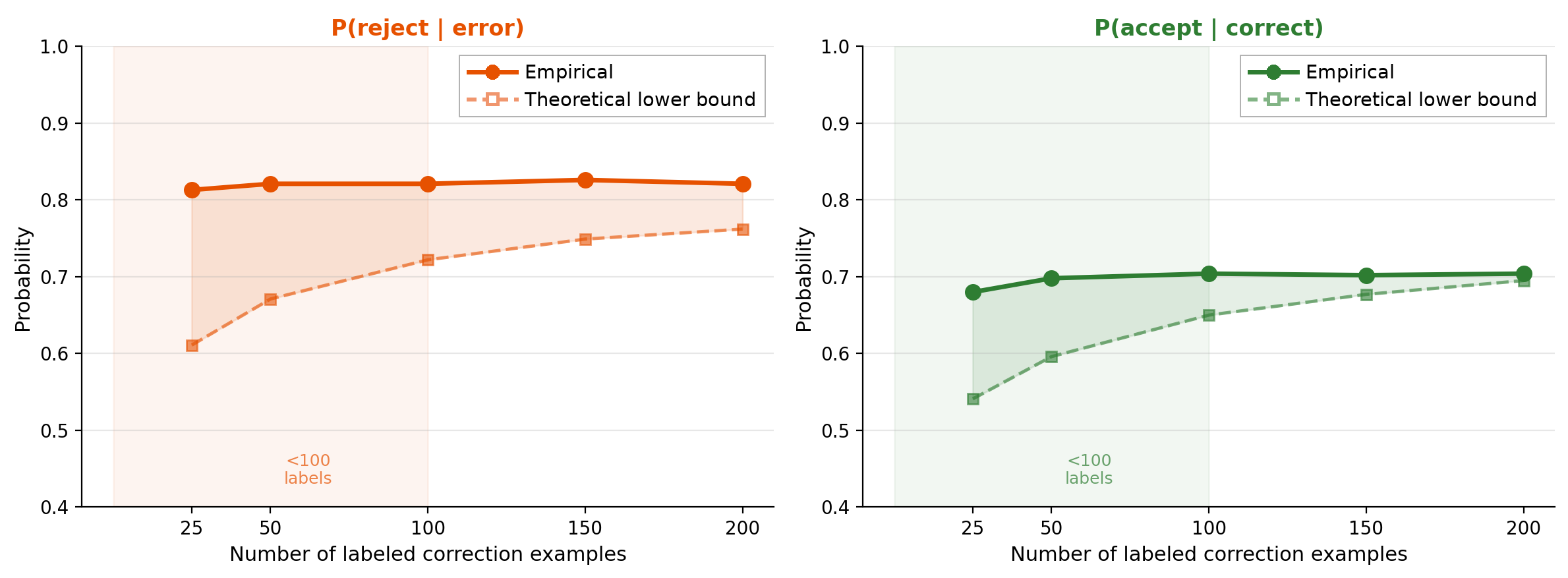}
\caption{Setup~1 corrector versus labelled correction budget.
Empirical performance is stable from 25 examples; theoretical bounds
tighten with larger calibration samples.}
\label{fig:h2_stress}
\end{figure}

All twenty empirical-versus-theoretical comparisons for the Transformer
corrector satisfy their distribution-agnostic lower bounds. With the synolitic
GNN score as $\Phi$, empirical error rejection is $0.821$ against a lower bound
of $0.722$. Confidence reaches $0.826$ and the graph-summary Fisher
representation $0.832$, showing that the guarantee is not specific to the graph
score. Performance remains stable with as few as $25$ labelled correction
examples; increasing calibration size mainly tightens the certificate
(Figure~\ref{fig:h2_stress}).

\begin{table}[!tbh]
\centering
\caption{Setup~1 Fisher corrector at $\Delta=0.8$.}
\label{tab:corrector}
\small
\begin{tabularx}{\linewidth}{@{}Lcccc@{}}
\toprule
Feature $\Phi$ & Accept bound & Accept empirical & Reject bound & Reject empirical\\
\midrule
GNN score & 0.650 & 0.704 & 0.722 & 0.821\\
Confidence & 0.512 & 0.559 & 0.722 & 0.826\\
Raw $X$ (Fisher) & 0.350 & 0.436 & 0.692 & 0.792\\
Graph topology (Fisher) & 0.279 & 0.332 & 0.692 & 0.792\\
Graph summary (Fisher) & 0.151 & 0.269 & 0.692 & 0.832\\
\bottomrule
\end{tabularx}
\end{table}

The near-ceiling fMRI decoder illustrates the opposite regime: test accuracy is
$0.9784$, but the calibration split contains only one error among $232$ graphs
(accuracy $0.9957$). With $M_-=1$ and $\Delta=0.8$, the theoretical rejection
bound is $\rho\approx0.025$. The guarantee becomes nearly vacuous because the
calibration sample contains too little error evidence, correctly refusing to
assert strong selective reliability.

\subsubsection{Dependence on labelled correction budget}
\label{sec:budget}

We evaluate how empirical correction performance and its theoretical
certificate change as the labelled calibration budget is reduced.

\subsubsection{Accept/reject trade-off}
\label{sec:tradeoff}

We additionally vary the confidence parameter $\Delta$ to characterize
the empirical accept/reject trade-off and the corresponding
distribution-agnostic lower bounds.

\begin{figure}[H]
\centering
\includegraphics[width=0.94\linewidth]{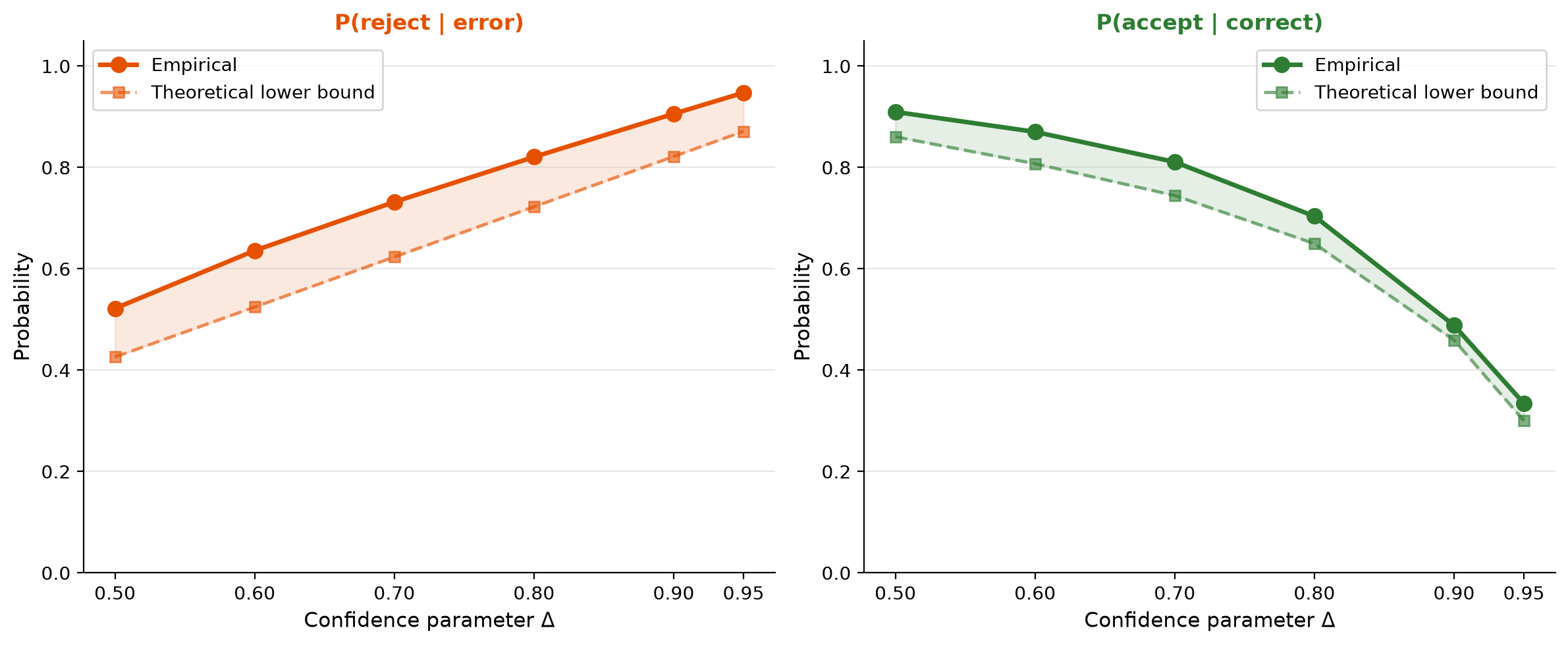}
\caption{Setup~1 corrector trade-off across confidence parameter
$\Delta$. Empirical accept/reject rates consistently satisfy the
corresponding lower bounds.}
\label{fig:h2_delta}
\end{figure}

\subsection{Cross-attention mechanism}
\label{sec:mechanism}
An initial analysis correlated synolitic node centrality with attention-head
specialization and found strong Spearman correlations ($|r|$ up to $0.935$).
These associations do not survive stratification by sequence length: attention
entropy is bounded by $\log L$ and rises with length, while node centrality
derived from $X$ falls with length, so the two can covary mechanically. This
motivated the direct cross-attention analysis reported below.

Direct cross-attention analysis yields a separate positive result. All four
layer-2 heads preferentially attend to the mirror source position (hit rates
$0.79$--$0.86$ versus chance near $0.07$), while layer-1 heads show no comparable
specialization. Within fixed length, correct predictions have greater mirror
fidelity than errors: sample-weighted gap $+0.091$, stratified-bootstrap $95\%$
CI $[+0.079,+0.104]$. Yet $140$ of $554$ errors occur despite intact mirror
alignment, showing that the reversal circuit can remain intact while decoding
fails downstream. EOS-timing errors account for approximately $50$--$90\%$ of
errors depending on sequence length. At $L=20$, per-token content accuracy
drops from $0.987$ to $0.941$; combining the measured termination failure with
conditional copy success reproduces the observed exact-match accuracy $0.377$.
This decomposition separates the redundant reversal circuit from downstream
EOS-timing and content-copy failures.

\section{Discussion}
The two regimes reveal different limits of graph-based interpretation. In the
Transformer, graph-derived scores inherit a recoverable nuisance variable; in
HCP, supervised edge sign already exposes the decisive target information
before higher-order processing begins. Thus graph construction can be useful
without implying that topology is the source of the predictive gain.

The HCP mechanism is particularly informative. Magnitude-only information is
near chance, so centralities built mainly from magnitude operate on a weak
channel. Signed node aggregation, by contrast, ensembles supervised edge
predictions. The small, consistently positive sign-separated-degree effect
suggests at most a weak local sign-structure contribution, not a broad
higher-order topology gain.

These conclusions are deliberately bounded. We evaluate one Transformer task,
one HCP contrast, one published pairwise edge construction, and one tuned GCN
family; other message-passing architectures or graph-generation schemes may
behave differently. The raw-HCP audit and published Vlasenko bundle are
distinct data paths, so no preprocessing guarantee is transferred between them.

The practical implication is an attribution discipline: preserve strong local
baselines and plausible nuisance variables before assigning credit to graph
topology. The Fisher corrector and mechanistic attention analysis are separate
positive results, not evidence for the topology claim.

\section{Conclusion}
Supervised pairwise graphs can organize useful predictive information, but
graph-level accuracy alone is insufficient evidence for higher-order topology.
In the Transformer setting, pairwise construction outperforms
attention-as-adjacency while its apparent advantage over confidence is explained
by sequence length. In HCP Emotion, signed local aggregation already captures
nearly all measurable signal available to the tested topology-aware summaries
and GCN.

The same study also shows that the representation can support provable post-hoc
correction and that direct cross-attention analysis reveals a robust reversal
circuit with downstream failure modes. The central methodological lesson is to
credit topology only after simpler local evidence and recoverable nuisance
variables have been excluded.

\section*{Author Contributions}
All authors contributed to the common experimental framework and interpretation.

\textbf{Ilya Brekin} designed and trained the controlled sequence-reversal
Transformer and developed shared reproducibility and data-interface
infrastructure. He led the sequence-length nuisance analysis and the mechanistic
cross-attention/error decomposition, and contributed to the HCP recording-length
and topology diagnostics.

\textbf{Ariela Moldavskaya} developed the supervised pairwise graph construction
for the Transformer, including the direction-invariant decisiveness
representation, and implemented graph-generation and contract-validation
infrastructure. She also implemented the raw-attention graph baseline and
developed the independent subject-safe raw-HCP path from time series through
out-of-fold pairwise graph generation to topology-aware PyTorch Geometric
objects.

\textbf{Elina Sidorova} developed and evaluated the graph-neural-network models
in both settings, including GATv2 for Transformer graphs and the
identity-preserving three-layer GCN for HCP ensemble graphs. She performed
validation-based architecture selection and the repeated-region
dimensional-scaling experiments and contributed to comparison with simpler
node-level representations.

\textbf{Ural Valeev} developed and implemented the Fisher post-hoc correction
component and its calibration procedure. He carried out the theoretical and
empirical evaluation of the distribution-agnostic guarantees across feature
representations, operating thresholds, and reduced labelled correction budgets.

The authors jointly designed the sequence of controls separating edge-level,
node-level, nuisance-derived, and higher-order information, interpreted the
combined results, and contributed to preparation and revision of the manuscript.

\section*{Acknowledgments}
We thank Professor Alexey Zaikin (University College London) for formulating
the research problem. We also thank the organizers and mentors of SMILES Summer
School 2026 for providing the project framework and research environment.
Data were provided in part by
the Human Connectome Project, WU-Minn Consortium (Principal Investigators:
David Van Essen and Kamil U\u{g}urbil; 1U54MH091657) funded by the 16 NIH Institutes
and Centers that support the NIH Blueprint for Neuroscience Research; and by the
McDonnell Center for Systems Neuroscience at Washington University.

\end{document}